\documentclass[11pt,a4paper]{article}
\usepackage[hyperref]{emnlp-ijcnlp-2019}
\usepackage{times}
\usepackage{latexsym}

\usepackage{graphicx}
\usepackage{float}
\usepackage{amsmath}
\usepackage{amssymb}
\usepackage{multirow}
\usepackage{booktabs}
\usepackage{url}
\usepackage{array}
\usepackage{enumitem}
\usepackage{algorithm}
\usepackage{algorithmic}
\usepackage{pifont}

\aclfinalcopy % Uncomment this line for the final submission

\title{Cross-Lingual Machine Reading Comprehension}

\author{Yiming Cui$^{\dag\ddag}$, Wanxiang Che$^\dag$, Ting Liu$^\dag$, Bing Qin$^\dag$, Shijin Wang$^\ddag$$^\S$, Guoping Hu$^\ddag$\\
{$^\dag$Research Center for Social Computing and Information Retrieval (SCIR),}\\
{Harbin Institute of Technology, Harbin, China}\\
{$^\ddag$State Key Laboratory of Cognitive Intelligence, iFLYTEK Research, China}\\
{$^\S$iFLYTEK AI Research (Hebei), Langfang, China} \\
{$^\dag$\tt \{ymcui,car,tliu,qinb\}@ir.hit.edu.cn}\\
{$^\ddag$$^\S$\tt\{ymcui,sjwang3,gphu\}@iflytek.com}\\  
}

\date{}

\begin{document}
\maketitle
\begin{abstract}
Though the community has made great progress on Machine Reading Comprehension (MRC) task, most of the previous works are solving English-based MRC problems, and there are few efforts on other languages mainly due to the lack of large-scale training data.
In this paper, we propose Cross-Lingual Machine Reading Comprehension (CLMRC) task for the languages other than English.
Firstly, we present several back-translation approaches for CLMRC task, which is straightforward to adopt. 
However, to accurately align the answer into another language is difficult and could introduce additional noise.
In this context, we propose a novel model called Dual BERT, which takes advantage of the large-scale training data provided by rich-resource language (such as English) and learn the semantic relations between the passage and question in a bilingual context, and then utilize the learned knowledge to improve reading comprehension performance of low-resource language.
We conduct experiments on two Chinese machine reading comprehension datasets CMRC 2018 and DRCD.
The results show consistent and significant improvements over various state-of-the-art systems by a large margin, which demonstrate the potentials in CLMRC task. 
\footnote{Resources available: \url{https://github.com/ymcui/Cross-Lingual-MRC}.}
\end{abstract}

%%%%%%%%%%%%%%%%%%%%%%
\section{Introduction}\label{introduction}
Machine Reading Comprehension (MRC) has been a popular task to test the reading ability of the machine, which requires to read text material and answer the questions based on it. Starting from cloze-style reading comprehension, various neural network approaches have been proposed and massive progresses have been made in creating large-scale datasets and neural models \citep{hermann-etal-2015,hill-etal-2015,kadlec-etal-2016,cui-acl2017-aoa,rajpurkar-etal-2016,dhingra-etal-2017}.
Though various types of contributions had been made, most works are dealing with English reading comprehension.
Reading comprehension in other than English has not been well-addressed mainly due to the lack of large-scale training data.

To enrich the training data, there are two traditional approaches.
Firstly, we can annotate data by human experts, which is ideal and high-quality, while it is time-consuming and rather expensive.
One can also obtain large-scale automatically generated data \citep{hermann-etal-2015,hill-etal-2015,liu-etal-2017}, but the quality is far beyond the usable threshold.
Another way is to exploit cross-lingual approaches to utilize the data in rich-resource language to implicitly learn the relations between $<$passage, question, answer$>$.

In this paper, we propose the Cross-Lingual Machine Reading Comprehension (CLMRC) task that aims to help reading comprehension in low-resource languages. 
First, we present several back-translation approaches when there is no or partially available resources in the target language.
Then we propose a novel model called Dual BERT to further improve the system performance when there is training data available in the target language.
We first translate target language training data into English to form pseudo bilingual parallel data. Then we use multilingual BERT \citep{devlin2018bert} to simultaneously model the $<$passage, question, answer$>$ in both languages, and fuse the representations of both to generate final predictions.
Experimental results on two Chinese reading comprehension dataset CMRC 2018 \citep{cui-emnlp2019-cmrc2018} and DRCD \citep{shao2018drcd} show that by utilizing English resources could substantially improve system performance and the proposed Dual BERT achieves state-of-the-art performances on both datasets, and even surpass human performance on some metrics. 
Also, we conduct experiments on the Japanese and French SQuAD \citep{asai2018multilingual} and achieves substantial improvements.
Moreover, detailed ablations and analysis are carried out to demonstrate the effectiveness of exploiting knowledge from rich-resource language.
To best of our knowledge, this is the first time that the cross-lingual approaches applied and evaluated on realistic reading comprehension data.
The main contributions of our paper can be concluded as follows.
\begin{itemize}[leftmargin=*]
  \item We present several back-translation based reading comprehension approaches and yield state-of-the-art performances on several reading comprehension datasets, including Chinese, Japanese, and French.
  \item We propose a model called Dual BERT to simultaneously model the $<$passage, question$>$ in both source and target language to enrich the text representations.
  \item Experimental results on two public Chinese reading comprehension datasets show that the proposed cross-lingual approaches yield significant improvements over various baseline systems and set new state-of-the-art performances.
\end{itemize}

%%%%%%%%%%%%%%%%%%%%%%%%%%%%%%%%%%%%%%%%%
\section{Related Works}\label{related-works}
Machine Reading Comprehension (MRC) has been a trending research topic in recent years. 
Among various types of MRC tasks, span-extraction reading comprehension has been enormously popular (such as SQuAD \citep{rajpurkar-etal-2016}), and we have seen a great progress on related neural network approaches \citep{wang-and-jiang-2016,seo-etal-2016,xiong-etal-2016,cui-acl2017-aoa,hu-etal-2018-read}, especially those were built on pre-trained language models, such as BERT \citep{devlin2018bert}.
While massive achievements have been made by the community, reading comprehension in other than English has not been well-studied mainly due to the lack of large-scale training data. 

\citet{asai2018multilingual} proposed to use runtime machine translation for multilingual extractive reading comprehension. 
They first translate the data from the target language to English and then obtain an answer using an English reading comprehension model. 
Finally, they recover the corresponding answer in the original language using soft-alignment attention scores from the NMT model. 
However, though an interesting attempt has been made, the zero-shot results are quite low, and alignments between different languages, especially for those have different word orders, are significantly different. 
Also, they only evaluate on a rather small dataset (hundreds of samples) that was translated from SQuAD \citep{rajpurkar-etal-2016}, which is not that realistic.

To solve the issues above and better exploit large-scale rich-resourced reading comprehension data, in this paper, we propose several zero-shot approaches which yield state-of-the-art performances on Japanese and French SQuAD data. 
Moreover, we also propose a supervised approach for the condition that there are training samples available for the target language.
To evaluate the effectiveness of our approach, we carried out experiments on two realistic public Chinese reading comprehension data: CMRC 2018 (simplified Chinese) \citep{cui-emnlp2019-cmrc2018} and DRCD (traditional Chinese) \citep{shao2018drcd}. 
Experimental results demonstrate the effectiveness by modeling training samples in a bilingual environment.

%%%%%%%%%%%%%%%%%%%%%%
\section{Back-Translation Approaches}\label{mta}
In this section, we illustrate back-translation approaches for cross-lingual machine reading comprehension, which is natural and easy to implement.
\begin{figure*}[tbp]
  \centering
  \includegraphics[width=1\textwidth]{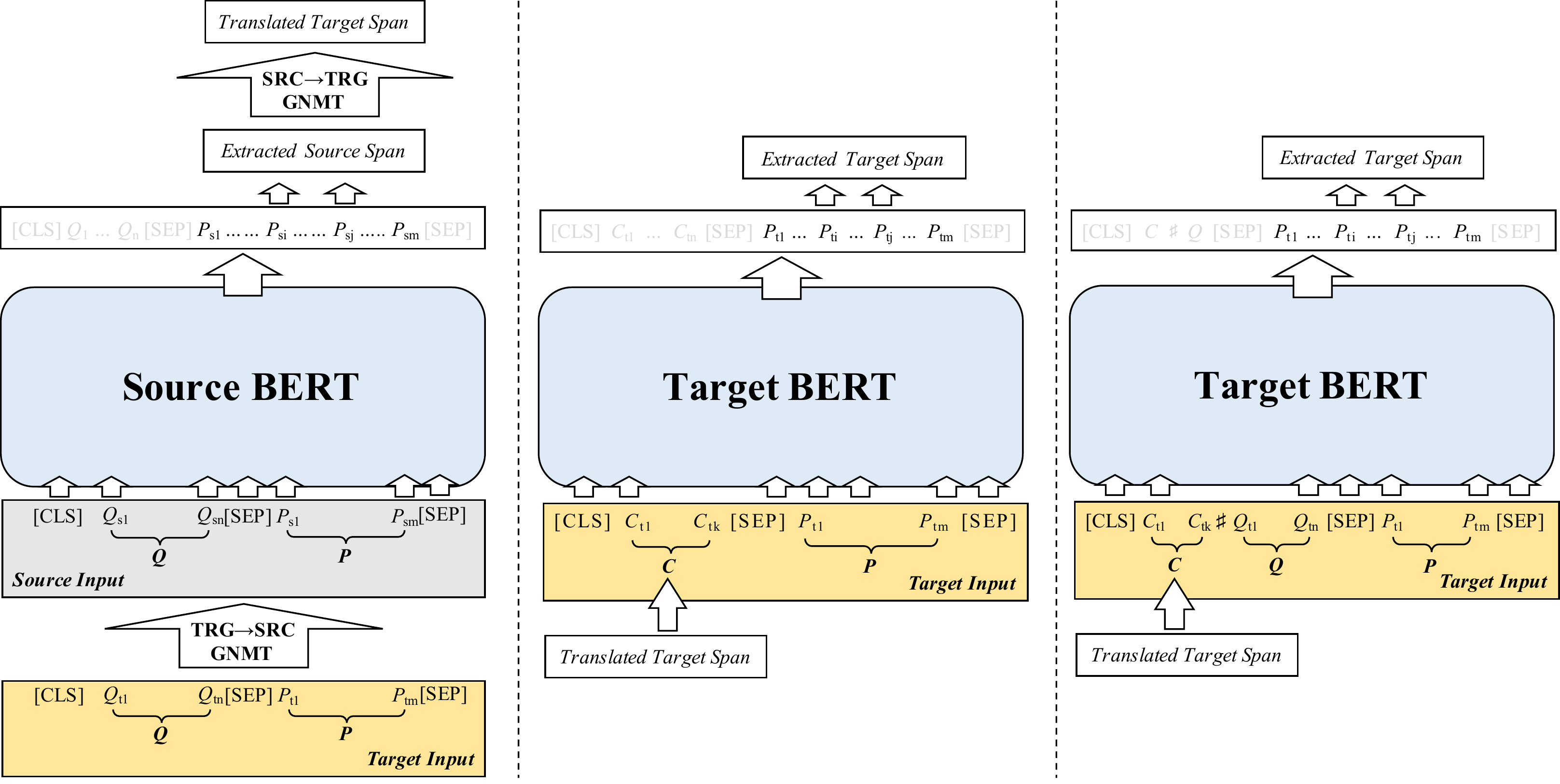}
  \caption{\label{nn-arch-gnmt} Back-translation approaches for cross-lingual machine reading comprehension (Left: GNMT, Middle: Answer Aligner, Right: Answer Verifier)}
\end{figure*}
Before introducing these approaches in detail, we will clarify crucial terminologies in this paper for better understanding. 
\begin{itemize}[leftmargin=*]
	\item {\bf Source Language}: Rich-resourced and has sufficient large-scale training data that we aim to extract knowledge from. We use subscript {\em S} for variables in the source language.
	\item {\bf Target Language}: Low-resourced and has only a few training data that we wish to optimize on. We use subscript {\em T} for variables in the target language.
\end{itemize}

In this paper, we aim to improve the machine reading comprehension performance in Chinese (target language) by introducing English (source language) resources.
The general idea of back-translation approaches is to translate $<$passage, question$>$ pair into the source language and generate an answer using a reading comprehension system in the source language. Finally, the generated answer is back-translated into the target language. 
In the following subsections, we will introduce several back-translation approaches for cross-lingual machine reading comprehension task. 
The architectures of the proposed back-translation approaches are depicted in Figure \ref{nn-arch-gnmt}.

%%%%%%%%%%%%%%%%%
\subsection{GNMT}
To build a simple cross-lingual machine reading comprehension system, it is straightforward to utilize translation system to bridge source and target language \citep{asai2018multilingual}.
Briefly, we first translate the target sample to the source language. 
Then we use a source reading comprehension system, such as BERT \citep{devlin2018bert}, to generate an answer in the source language. 
Finally, we use back-translation to get the answer in the target language. 
As we do not exploit any training data in the target language, we could regard this approach as a {\em zero-shot} cross-lingual baseline system.

Specifically, we use Google Neural Machine Translation (GNMT) system for source-to-target and target-to-source translations.
One may also use advanced and domain-specific neural machine translation system to achieve better translation performance, while we leave it for individuals, and this is beyond the scope of this paper.

However, for span-extraction reading comprehension task, a major drawback of this approach is that the translated answer may not be the exact span in the target passage.
To remedy this, we propose three simple approaches to improve the quality of the translated answer in the target language.

%%%%%%%%%%%%%%%%%
\subsection{Simple Match}
We propose a simple approach to align the translated answer into extract span in the target passage.
We calculate character-level text overlap (for Chinese) between translated answer $A_{trans}$ and arbitrary sliding window in target passage $\mathcal{P}_{T[i:j]}$.
The length of sliding window ranges $len(A_{trans}) \pm \delta$, with a relax parameter $\delta$. Typically, the relax parameter $\delta \in [0,5]$ as the length between ground truth and translated answer does not differ much in length.
In this way, we would calculate character-level F1-score of each candidate span $\mathcal{P}_{T[i:j]}$ and translated answer $A_{trans}$, and we could choose the best matching one accordingly.
Using the proposed SimpleMatch could ensure the predicted answer is an exact span in target passage.
As SimpleMatch does not use target training data either, it could also be a pipeline component in {\em zero-shot} settings.

%%%%%%%%%%%%%%%%%
\subsection{Answer Aligner}
Though we could use unsupervised approaches for aligning answer, such as the proposed SimpleMatch, it stops at token-level and lacks  semantic awareness between the translated answer and ground truth answer.
In this paper, we also propose two supervised approaches for further improving the answer span when there is training data available in the target language.

The first one is Answer Aligner, where we feed translated answer $\mathcal{A}_{trans}$ and target passage $\mathcal{P}_{T}$ into the BERT and outputs the ground truth answer span $\mathcal{A}_{T}$. 
The model will learn the semantic relations between them and generate improved span for the target language.

\begin{figure*}[tbp]
  \centering
  \includegraphics[width=0.9\textwidth]{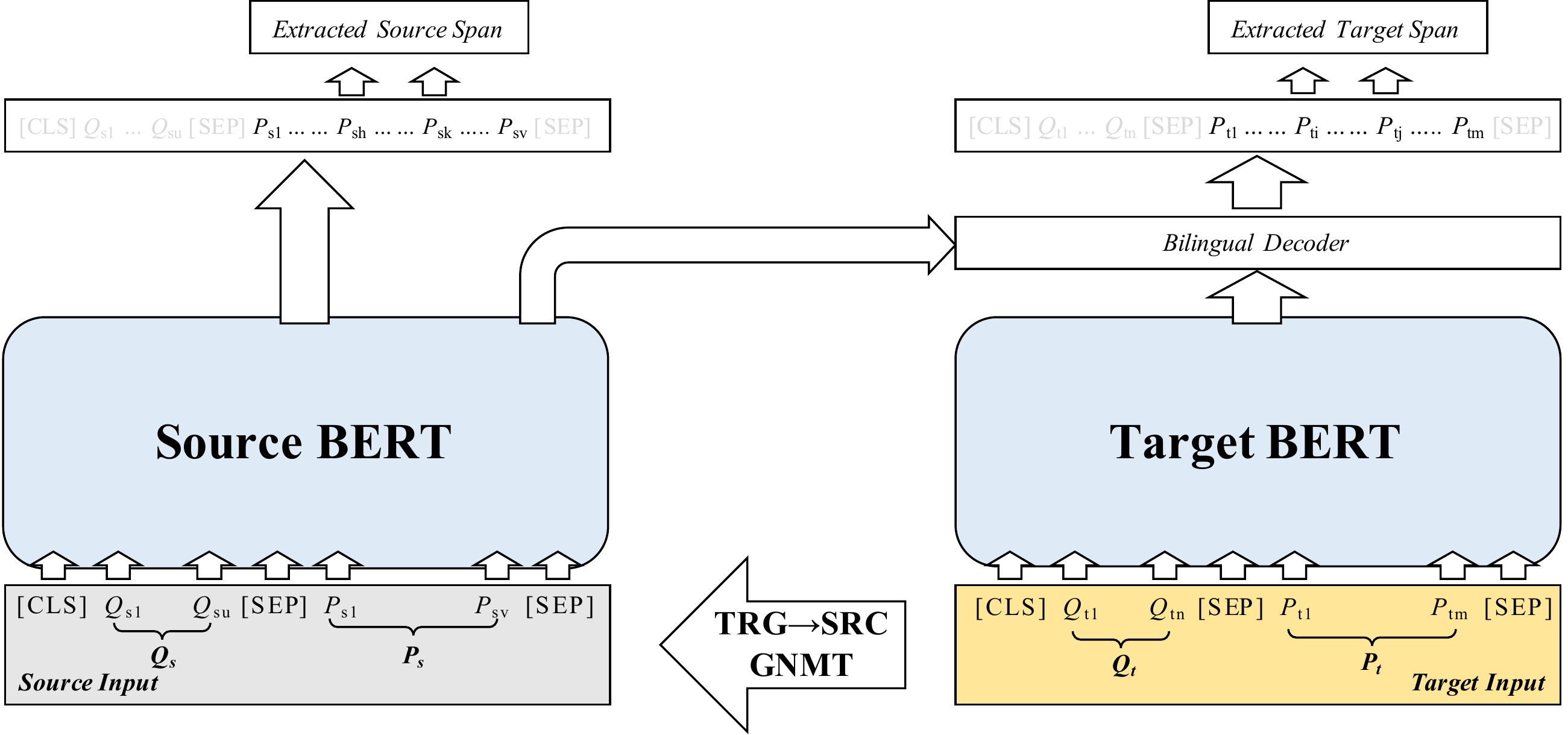}
  \caption{\label{nn-arch} System overview of the Dual BERT model for cross-lingual machine reading comprehension task.}
\end{figure*}

%%%%%%%%%%%%%%%%%
\subsection{Answer Verifier}
In Answer Aligner, we did not exploit question information in target training data.
One can also utilize question information to transform Answer Aligner into Answer Verifier, as we use complete $\langle \mathcal{P}_T, \mathcal{Q}_T, \mathcal{A}_T \rangle$ in the target language and additional translated answer $\mathcal{A}_{trans}$ to verify its correctness and generate improved span.

%%%%%%%%%%%%%%%%%%%%%%%%%%%%%
\section{Dual BERT}\label{dual-bert}
One disadvantage of the back-translation approaches is that we have to recover the source answer into the target language.
To remedy the issue, in this paper, we propose a novel model called Dual BERT to simultaneously model the training data in both source and target language to better exploit the relations among $<$passage, question, answer$>$. 
The model could be used when there is training data available for the target language, and we could better utilize source language data to enhance the target reading comprehension system. 
The overall neural architecture for Dual BERT is shown in Figure \ref{nn-arch}.

\subsection{Dual Encoder}
Bidirectional Encoder Representation from Transformers (BERT) has shown marvelous performance in various NLP tasks, which substantially outperforms non-pretrained models by a large margin \citep{devlin2018bert}.
In this paper, we use multi-lingual BERT for better encoding the text in both source and target language.
Formally, given target passage $\mathcal{P}_{T}$ and question $\mathcal{Q}_{T}$, we organize the input $X_{T}$ for BERT as follows.
\begin{quote}
	%\small
	{\tt [CLS] ${\mathcal{Q}_{T}}$ [SEP] ${\mathcal{P}_{T}}$ [SEP]}
\end{quote}
Similarly, we can also obtain source training sample by translating target sample with GNMT, forming input $X_{S}$ for BERT.
Then we use $X_{T}$ and $X_{S}$ to obtain deep contextualized representations through a {\em shared} multi-lingual BERT, forming $B_{T}\in\mathbb{R}^{L_T*h}, B_{S}\in\mathbb{R}^{L_S*h}$, where $L$ represents the length of input and $h$ is the hidden size (768 for multi-lingual BERT).

%%%%%%%%%%%%%%%%%
\subsection{Bilingual Decoder}
Typically, in the reading comprehension task, attention mechanism is used to measure the relations between the passage and question.
Moreover, as Transformers are fundamental components of BERT, multi-head self-attention layer \citep{vaswani2017attention} is used to extract useful information within the input sequence.

Specifically, in our model, to enhance the target representation, we use a multi-head self-attention layer to extract useful information in source BERT representation $B_{S}$.
We aim to generate target span by not only relying on target representation but also on source representation to simultaneously consider the $<$passage, question$>$ relations in both languages, which can be seen as a bilingual decoding process.

Briefly, we regard target BERT representation $B_T$ as {\em query} and source BERT representation $B_S$ as {\em key} and {\em value} in multi-head attention mechanism.
In original multi-head attention, we calculate a raw dot attention as follows. \footnote{We omit rather extensive formulations of representation transformations and kindly advise the readers refer to the attention implementation in BERT: https://github.com/google-research/bert/blob/master/modeling.py\#L558}
This will result in an attention matrix $A_{TS}$ that indicate raw relations between each source and target token.
\begin{gather} 
A_{TS} = B_T \cdot B_S^\top, ~~A_{TS} \in\mathbb{R}^{L_T*L_S}
\end{gather}

To combine the benefit of both inter-attention and self-attention, instead of using Equation 1, we propose a simple modification on multi-head attention mechanism, which is called {\bf Self-Adaptive Attention (SAA)}.
First, we calculate self-attention of $B_T$ and $B_S$ and apply the softmax function, as shown in Equation 2 and 3. 
This is designed to use self-attention to filter the irrelevant part within each representation firstly, and inform the raw dot attention on paying more attention to the self-attended part, making the attention more precise and accurate.
\begin{gather} 
A_T =\mathbf{softmax}(B_T \cdot B_T^\top) \\
A_S = \mathbf{softmax}(B_S \cdot B_S^\top)
\end{gather}
Then we use self-attention $A_T$ and $A_S$, inter-attention $A_{TS}$ to get self-attentive attention $\tilde{A}_{TS}$.
We calculate dot product between ${A}_{ST}$ and $B_S$ to obtain attended representation $R' \in \mathbb{R}^{L_T*h}$.
\begin{gather} 
\tilde{A}_{TS} = A_T \cdot A_{TS} \cdot {A_S}^\top , \tilde{A}_{TS} \in \mathbb{R}^{L_T*L_S} \\
R' = \mathbf{softmax}(\tilde{A}_{TS}) \cdot B_S 
\end{gather}

After obtaining attended representation $R'$, we use an additional fully connected layer with residual layer normalization which is similar to BERT implementation.
\begin{gather}
R = W_r R' + b_r,~~W_r \in\mathbb{R}^{h*h} \\
H_{T} = concat[B_T, \mathbf{LayerNorm}(B_T + R)]
\end{gather}

Finally, we calculate weighted sum of $H_{T}$ to get final span prediction $P_{T}^{s}, P_{T}^{e}$ (superscript $s$ for start, $e$ for end).
For example, the start position $P_{T}^{s}$ is calculated by the following equation.
\begin{gather} 
P_{T}^{s} = \mathbf{softmax}(W_{T}^\top H_{T} + b),~~W_{T}\in\mathbb{R}^{2h}
\end{gather}

We calculate standard cross entropy loss for the start and end predictions in the target language.
	\begin{equation}
		\mathcal{L}_{T} = -\frac{1}{N} \sum\limits_{i=1}^{N}(y_{T}^{s}\log(P_{T}^{s}) + y_{T}^{e}\log(P_{T}^{e}) )
	\end{equation}

%%%%%%%%%%%%%%%%%
\subsection{Auxiliary Output}
In order to evaluate how translated sample behaves in the source language system, we also generate span prediction for source language using BERT representation $B_S$ directly without further calculation, resulting in the start and target prediction $P_{S}^{s}, P_{S}^{e}$ (similar to Equation 8).
Moreover, we also calculate cross-entropy loss $\mathcal{L}_{aux}$ for translated sample (similar to Equation 9), where a $\lambda$ parameter is applied to this loss.

Instead of setting $\lambda$ with heuristic value, in this paper, we propose a novel approach to better adjust $\lambda$ automatically.
As the sample was generated by the machine translation system, there would be information loss during the translation process. 
Wrong or partially translated samples may harm the performance of reading comprehension system. 
To measure how the translated samples assemble the real target samples, we calculate cosine similarity between the {\em ground truth} span representation in source and target language (denoted as $\tilde{H}_{S}$ and $\tilde{H}_{T}$). 
When the ground truth span representation in the translated sample is similar to the real target samples, the $\lambda$ increase; otherwise, we only use target span loss as $\lambda$ may decrease to zero.

The span representation is the concatenation of three parts: BERT representation of ground truth start $B^s  \in \mathbb{R}^{h} $, ground truth end $B^e  \in \mathbb{R}^{h}$, and self-attended span $B^{att}  \in \mathbb{R}^{h}$, which considers both boundary information (start/end) and mixed representation of the whole ground truth span. 
We use BERT representation $B$\footnote{We mask out the values that out of span.} to get a self-attended span representation $B^{att}$ using a simple dot product with average pooling, to get a 2D-tensor.
\begin{gather}
	\tilde{H}_{S} = concat[B_{S}^{s}, B_{S}^{e}, B_{S}^{att}] \\
	\tilde{H}_{T} = concat[B_{T}^{s}, B_{T}^{e}, B_{T}^{att}] \\
	\lambda = \max \{0, \cos <\tilde{H}_{S}, \tilde{H}_{T}>\}
\end{gather}

The overall loss for Dual BERT is composed by two parts: target span loss $\mathcal{L}_{T}$ and auxiliary span loss in source language $\mathcal{L}_{aux}$.
\begin{gather} 
\mathcal{L} =  \mathcal{L}_{T} + \lambda \mathcal{L}_{aux}
\end{gather}

%%%%%%%%%%%%%%%%%%%%%%%%%%%%%%%%%%%%%%%%%
\section{Experiments}\label{experiments}
\subsection{Experimental Setups}
We evaluate our approaches on two public Chinese span-extraction machine reading comprehension datasets: CMRC 2018 (simplified Chinese) \citep{cui-emnlp2019-cmrc2018}\footnote{https://github.com/ymcui/cmrc2018/} and DRCD (traditional Chinese) \citep{shao2018drcd}\footnote{https://github.com/DRCSolutionService/DRCD/}. 
The statistics of the two datasets are listed in Table \ref{data-stats}. 
\begin{table}[htbp]
\small
\begin{center}
\begin{tabular}{l rrrr}
\toprule
 & \bf Train & \bf Dev & \bf Test & \bf Challenge \\
\midrule
\bf \em CMRC 2018 \\
Question \# & 10,321 & 3,219 & 4,895 & 504 \\
Answer \#   & 1 & 3 & 3 & 3 \\
\midrule
\bf \em DRCD \\
Question \# & 26,936 & 3,524 & 3,493 & - \\
Answer \#   & 1 & 2 & 2 & - \\
\bottomrule
\end{tabular}
\end{center}
\caption{\label{data-stats} Statistics of CMRC 2018 and DRCD.}
\end{table}

Note that, since the test and challenge sets are preserved by CMRC 2018 official to ensure the integrity of the evaluation process, we submitted our best-performing systems to the organizers to get these scores. 
The resource in source language was chosen as SQuAD \citep{rajpurkar-etal-2016} training data.
The settings of the proposed approaches are listed below in detail.
\begin{itemize}[leftmargin=*]
  \item {\bf Tokenization}: Following the official BERT implementation, we use WordPiece tokenizer \citep{wu2016google} for English and character-level tokenizer for Chinese.
  \item {\bf BERT}: We use pre-trained English BERT on SQuAD 1.1 \citep{rajpurkar-etal-2016} for initialization, denoted as SQ-$B_{en}$ (base) and SQ-$L_{en}$ (large) for back-translation approaches. For other conditions, we use multi-lingual BERT as default, denoted as $B_{mul}$ (and SQ-$B_{mul}$ for those were pre-trained on SQuAD).\footnote{https://github.com/google-research/bert}
  \item {\bf Translation}: We use Google Neural Machine Translation (GNMT) system for translation.\footnote{https://cloud.google.com/translate/} We evaluated GNMT system on NIST MT02/03/04/05/06/08 Chinese-English set and achieved an average BLEU score of 43.24, compared to previous best work (43.20) \citep{robust-nmt-acl2018}, yielding state-of-the-art performance. 
  \item {\bf Optimization}: Following original BERT implementation, we use \textsc{Adam} with weight decay optimizer \citep{kingma2014adam} using an initial learning rate of 4e-5 and use cosine learning rate decay scheme instead of the original linear decay, which we found it beneficial for stabilizing results. The training batch size is set to 64, and each model is trained for 2 epochs, which roughly takes 1 hour.
  \item {\bf Implementation}: We modified the TensorFlow \citep{abadi2016tensorflow} version {\tt run\_squad.py} provided by BERT. All models are trained on Cloud TPU v2 that has 64GB HBM.
\end{itemize}

        \begin{table*}[ht]
        \small
        \begin{center}
        \begin{tabular}{p{0.1cm}  p{4.8cm}  p{0.6cm}<{\centering} p{0.6cm}<{\centering} p{0.6cm}<{\centering} p{0.6cm}<{\centering}  p{0.6cm}<{\centering} p{0.6cm}<{\centering} |p{0.6cm}<{\centering} p{0.6cm}<{\centering} p{0.6cm}<{\centering} p{0.6cm}<{\centering}}
        \toprule
        \multirow{3}{*}{\bf \#} & \multirow{3}{*}{\bf System} & \multicolumn{6}{c|}{\centering \bf CMRC 2018} & \multicolumn{4}{c}{\centering \bf DRCD} \\
        & & \multicolumn{2}{c}{\centering \bf  Dev} & \multicolumn{2}{c}{\centering \bf Test} & \multicolumn{2}{c|}{\centering \bf Challenge} & \multicolumn{2}{c}{\centering \bf Dev} & \multicolumn{2}{c}{\centering \bf Test}\\
        & & {\bf EM} & {\bf F1} &{\bf EM} & {\bf F1} & {\bf EM} & {\bf F1} &{\bf EM} & {\bf F1} &{\bf EM} & {\bf F1} \\
        \midrule
        & {\em Human Performance}                 & {\em 91.1} & {\em 97.3} & {\em 92.4} & {\em 97.9} & {\em 90.4} & {\em 95.2} & - & - & {\em 80.4} & {\em 93.3} \\
        & P-Reader (single model)$^\dag$      	& 59.9 & 81.5 & 65.2 & 84.4 & 15.1 & 39.6 & - & - & - & - \\
        & Z-Reader (single model)$^\dag$       	& 79.8 & 92.7 & 74.2 & 88.1 & 13.9 & 37.4 & - & - & - & - \\
        & MCA-Reader (ensemble)$^\dag$       	& 66.7 & 85.5 & 71.2 & 88.1 & 15.5 & 37.1 & - & - & - & -  \\
        & RCEN (ensemble)$^\dag$         		& 76.3 & 91.4 & 68.7 & 85.8 & 15.3 & 34.5 & - & - & - & -  \\
        & r-net (single model)$^\dag$       		& - & - & - & - & - & - & - & - & 29.1 & 44.4 \\
        & DA \citep{yang-da-2019}			& 49.2 & 65.4 & - & - & - & - & 55.4 & 67.7 & - & - \\
        \midrule\midrule
        1& GNMT+BERT$_{SQ-B_{en}}$$^\spadesuit$          			& 15.9 & 40.3 & 20.8 & 45.4 & 4.2 & 20.2 & 28.1 & 50.0 & 26.6 & 48.9 \\
        2& GNMT+BERT$_{SQ-L_{en}}$$^\spadesuit$          			& 16.8 & 42.1 & 21.7 & 47.3 & 5.2 & 22.0 & 28.9 & 52.0 & 28.7 & 52.1 \\
        3& GNMT+BERT$_{SQ-L_{en}}$+SimpleMatch$^\spadesuit$   	& 26.7 & 56.9 & 31.3 & 61.6 & 9.1 & 35.5 & 36.9 & 60.6 & 37.0 & 61.2 \\
        4& GNMT+BERT$_{SQ-L_{en}}$+Aligner					& 46.1 & 66.4 & 49.8 & 69.3 & 16.5 & 40.9 & 60.1 & 70.5 & 59.5 & 70.7 \\
        5& GNMT+BERT$_{SQ-L_{en}}$+Verifier     				& 64.7 & 84.7 & 68.9 & 86.8 & 20.0 & 45.6 & 83.5 & 90.1 & 82.6 & 89.6 \\
        \midrule \midrule
        6& BERT$_{B_{cn}}$            							& 63.6 & 83.9 & 67.8 & 86.0 & 18.4 & 42.1 & 83.4 & 90.1 & 81.9 & 89.0 \\
        7& BERT$_{B_{mul}}$            							& 64.1 & 84.4 & 68.6 & 86.8 & 18.6 & 43.8 & 83.2 & 89.9 & 82.4 & 89.5 \\
        8& {\bf Dual BERT} 			 					& 65.8 & 86.3 & 70.4 & 88.1 & 23.8 & 47.9 & 84.5 & 90.8 & 83.7 & 90.3 \\
        \midrule
        9& BERT$_{SQ-B_{mul}}$$^\spadesuit$     				& 56.5 & 77.5 & 59.7 & 79.9 & 18.6 & 41.4 & 66.7 & 81.0 & 65.4 & 80.1 \\
        10& BERT$_{SQ-B_{mul}}$ + Cascade Training    			& 66.6 & 87.3 & 71.8 & 89.4 & 25.6 & 52.3 & 85.2 & 91.4 & 84.4 & 90.8 \\
        11& BERT$_{B_{mul}}$ + Mixed Training    				& 66.8 & 87.5 & 72.6 & 89.8 & 26.7 & 53.4 & 85.3 & 91.6 & 84.7 & 91.2 \\
        12& \bf Dual BERT (w/ SQuAD)      						& 68.0 & 88.1 & 73.6 & 90.2 & 27.8 & 55.2 & 86.0 & 92.1 & 85.4 & 91.6 \\
        \bottomrule
        \end{tabular}
        \end{center}
        \caption{\label{overall-results} Experimental results on CMRC 2018 and DRCD. $^\dag$ indicates unpublished works (some of the systems are using development set for training, which makes the results not directly comparable.). $^\spadesuit$ indicates zero-shot approach. We mark our system with an ID in the first column for reference simplicity. }
        \end{table*}

%%%%%%%%%%%%%%%%%%%%%%%%%%%%%%%%
\subsection{Overall Results}
The overall results are shown in Table \ref{overall-results}. 
As we can see that, without using any alignment approach, the zero-shot results are quite lower regardless of using English BERT-base (\#1) or BERT-large (\#2). When we apply SimpleMatch (\#3), we observe significant improvements demonstrating its effectiveness.
The Answer Aligner (\#4)  could further improve the performance beyond SimpleMatch approach, demonstrating that the machine learning approach could dynamically adjust the span output by learning the semantic relationship between translated answer and target passage. 
Also, the Answer Verifier (\#5) could further boost performance and surpass the multi-lingual BERT baseline (\#7) that only use target training data, demonstrating that it is beneficial to adopt rich-resourced language to improve machine reading comprehension in other languages.

When we do not use SQuAD pre-trained weights, the proposed Dual BERT (\#8) yields significant improvements (all results are verified by p-test with $p<0.05$) over both Chinese BERT (\#6) and multi-lingual BERT (\#7) by a large margin. 
If we only train the BERT with SQuAD (\#9), which is a zero-shot system, we can see that it achieves decent performance on two Chinese reading comprehension data. 
Moreover, we can also pursue further improvements by continue training (\#10) with Chinese data starting from the system \#9, or mixing Chinese data with SQuAD and training from initial multi-lingual BERT (\#11). 
Under powerful SQuAD pre-trained baselines, Dual BERT (\#12) still gives moderate and consistent improvements over Cascade Training (\#10) and Mixed Training (\#11) baselines and set new state-of-the-art performances on both datasets, demonstrating the effectiveness of using machine-translated sample to enhance the Chinese reading comprehension performance.

%%%%%%%%%%%%%%%%%%%%%%%%%%%%%%%%
\subsection{Results on Japanese and French SQuAD}
In this paper, we propose a simple but effective approach called SimpleMatch to align translated answer to original passage span.
While one may argue that using neural machine translation attention to project source answer to original target passage span is ideal as used in  \citet{asai2018multilingual}. 
However, to extract attention value in neural machine translation system and apply it to extract the original passage span is bothersome and computationally ineffective.
To demonstrate the effectiveness of using SimpleMatch instead of using NMT attention to extract original passage span in zero-shot condition, we applied SimpleMatch to Japanese and French SQuAD (304 samples for each) which is what exactly used in \citet{asai2018multilingual}. 
The results are listed in Table \ref{result-zero-shot}.

\begin{table}[tp]
\small
\begin{center}
\begin{tabular}{p{3cm} c c c c}
\toprule
& \multicolumn{2}{c}{\centering \bf Japanese} & \multicolumn{2}{c}{\centering \bf French} \\
& \bf EM & \bf F1 & \bf EM & \bf F1  \\
\midrule
Back-Translation$\dag$  	& 24.8 & 42.6 & 23.5 & 44.0 \\
~~+Runtime MT$\dag$    		& 37.0 & 52.2 & 40.7 & 61.9 \\
\midrule
GNMT+BERT$_{L_{en}}$  		& 26.9 & 46.2 & 39.1 & 67.0 \\
~~+SimpleMatch    			& 37.3 & 58.0 & 47.4 & 71.5 \\
\midrule
BERT$_{SQ-B_{mul}}$ 			& 61.3 & 73.4 & 57.6 & 77.1 \\
\bottomrule
\end{tabular}
\end{center}
\caption{\label{result-zero-shot} Zero-shot cross-lingual machine reading comprehension results on Japanese and French SQuAD data. $\dag$ are extracted in \citet{asai2018multilingual}.}
\end{table}

From the results, we can see that, though our baseline (GNMT+BERT$_{L_{en}}$) is higher than previous work (Back-Translation \citep{asai2018multilingual}), when using SimpleMatch to extract original passage span could obtain competitive of even larger improvements.
In Japanese SQuAD, the F1 score improved by 9.6 in \citet{asai2018multilingual} using NMT attention, while we obtain larger improvement with 11.8 points demonstrating the effectiveness of the proposed method.
BERT with pre-trained SQuAD weights yields the best performance among these systems, as it does not require the machine translation process and has unified text representations for different languages.

%%%%%%%%%%%%%%%%
\subsection{Ablation Studies}
In this section, we ablate important components in our model to explicitly demonstrate its effectiveness.
The ablation results are depicted in Table \ref{result-ablation}.

\begin{table}[ht]
\small
\begin{center}
\begin{tabular}{p{3.5cm} ll}
\toprule
& \bf EM & \bf F1 \\
\midrule
\bf Dual BERT (w/ SQuAD)    		& \bf 68.0 & \bf 88.1 \\
~~w/o Auxiliary Loss				& 67.5 (-0.5)  & 87.7 (-0.4)  \\
~~w/o Dynamic Lambda			& 67.3 (-0.7)  & 87.5 (-0.6)  \\
~~w/o Self-Adaptive Att.			& 67.2 (-0.8)  & 87.5 (-0.6)  \\
~~w/o Source BERT				& 66.6 (-1.4) & 87.3 (-0.8) \\
~~w/o SQuAD Pre-Train	 		& 65.8 (-2.2)  & 86.3 (-1.8) \\
\bottomrule
\end{tabular}
\end{center}
\caption{\label{result-ablation} Ablations of Dual BERT on the CMRC 2018 development set.}
\end{table}

As we can see that, removing SQuAD pre-trained weights (i.e., using randomly initialized BERT) hurts the performance most, suggesting that it is beneficial to use pre-trained weights though the source and the target language is different.
Removing source BERT will degenerate to cascade training, and the results show that it also harms overall performance, demonstrating that it is beneficial to utilize translated sample for better characterizing the relations between $<$passage, question, answer$>$.
The other modifications seem to also consistently decrease the performance to some extent, but not as salient as the data-related components (last two lines), indicating that data-related approaches are important in cross-lingual machine reading comprehension task.

%%%%%%%%%%%%%%%%%%%%%%%%%%%%%%%%
\section{Discussion}
In our preliminary cross-lingual experiments, we adopt English as our source language data.
However, one question remains unclear.

{\em Is it better to pre-train with larger data in a distant language (such as English, as oppose to Simplified Chinese), or with smaller data in closer language (such as Traditional Chinese)?}

\begin{figure}[tbp]
  \centering
  \includegraphics[width=0.48\textwidth]{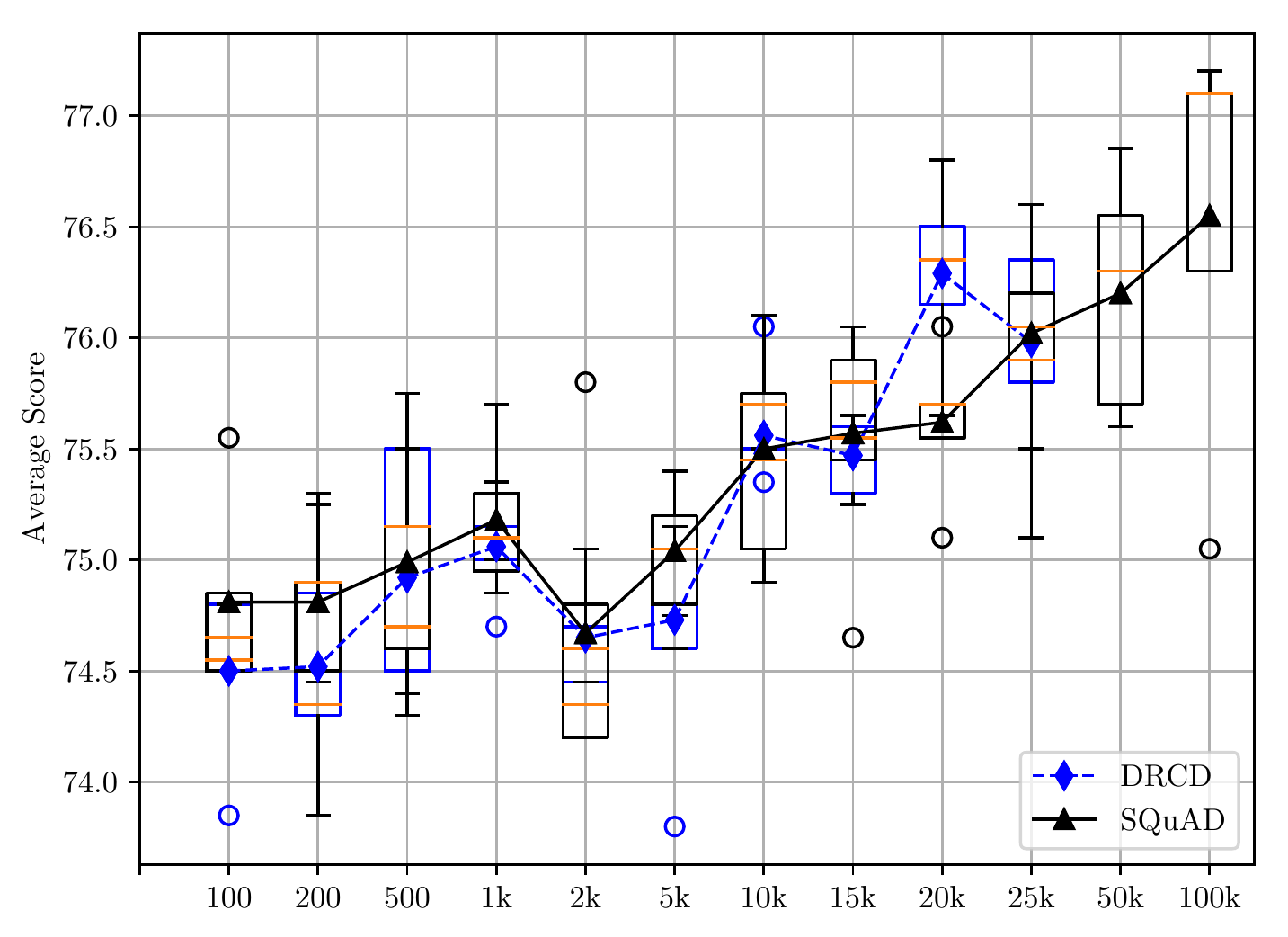}
  \caption{\label{analysis} BERT performance (average of EM and F1) with different amount of pre-training SQuAD (English) or DRCD (Traditional Chinese). }
\end{figure}

To investigate the problem, we plot the multi-lingual BERT performance on the CMRC 2018 development data using different language and data size in the pre-training stage.
The results are depicted in Figure \ref{analysis}, and we come to several observations.

Firstly, when the size of pre-training data is under 25k (training data size of DRCD), we can see that there is no much difference whether we use Chinese or English data for pre-training, and even the English pre-trained models are better than Chinese pre-trained models in most of the times, which is not expected. 
We suspect that, by using multi-lingual BERT, the model tend to provide universal representations for the text and learn the language-independent semantic relations among the inputs which is ideal for cross-lingual tasks, thus the model is not that sensitive to the language in the pre-training stage.
Also, as training data size of SQuAD is larger than DRCD, we could use more data for pre-training. 
When we add more SQuAD data ($>$25k) in the pre-training stage, the performance on the downstream task (CMRC 2018) continues to improve significantly. 
In this context, we conclude that,
\begin{itemize}
	\item When the pre-training data is not abundant, there is no special preference on the selection of source (pre-training) language.
	\item If there are large-scale training data available for several languages, we should select the source language as the one that has the largest training data rather than its linguistic similarity to the target language. 
\end{itemize}

Furthermore, one could also take advantages of data in various languages, but not only in a bilingual environment, to further exploit knowledge from various sources, which is beyond the scope of this paper and we leave this for future work.

%%%%%%%%%%%%%%%%%%%%%%%%%%%%%%%%%%%%%%%%%
\section{Conclusion}\label{conclusion}
In this paper, we propose Cross-Lingual Machine Reading Comprehension (CLMRC) task.
When there is no training data available for the target language, firstly, we provide several zero-shot approaches that were initially trained on English and transfer to other languages, along with three methods to improve the translated answer span by using unsupervised and supervised approaches.
When there is training data available for the target language, we propose a novel model called Dual BERT to simultaneously model the $<$passage, question, answer$>$ in source and target languages using multi-lingual BERT.
The proposed method takes advantage of the large-scale training data by rich-resource language (such as SQuAD) and learns the semantic relations between the passage and question in both source and target language.
Experiments on two Chinese machine reading comprehension datasets indicate that the proposed model could give consistent and significant improvements over various state-of-the-art systems by a large margin and set baselines for future research on CLMRC task.

Future studies on cross-lingual machine reading comprehension will focus on 
1) how to utilize various types of English reading comprehension data; 
2) cross-lingual machine reading comprehension without the translation process, etc.

%%%%%%%%%%%%%%%%%%%%%%%%%%%%%%%%%%%%%%%%
\section*{Acknowledgments}
We would like to thank all anonymous reviewers for their thorough reviewing and providing constructive comments to improve our paper. 
The first author was partially supported by the Google TensorFlow Research Cloud (TFRC) program for Cloud TPU access.
This work was supported by the National Natural Science Foundation of China (NSFC) via grant 61976072, 61632011, and 61772153.

\bibliography{emnlp-ijcnlp-2019}
\bibliographystyle{acl_natbib}

\end{document}